\documentclass{ecai} 

\usepackage{latexsym}
\usepackage{amssymb}
\usepackage{amsmath}
\usepackage{amsthm}
\usepackage{booktabs}
\usepackage{enumitem}
\usepackage{graphicx}
\usepackage{color}
\usepackage{xspace}
\usepackage{algorithm}
\usepackage{algorithmicx}
\usepackage{algpseudocode}
\usepackage{multirow}
\usepackage[most]{tcolorbox}  
\usepackage{tcolorbox}
\usepackage{footmisc}

\definecolor{darkgrey}{RGB}{120,120,120}
\definecolor{mygrey}{RGB}{200,200,200}
\newcommand{\name}[0]{LGMformer\xspace}

\renewcommand{\thefootnote}{\fnsymbol{footnote}}

\newtheorem{observation}{Observation}

\begin{document}
\begin{frontmatter}
\title{Learning a Mini-batch Graph Transformer via Two-stage Interaction Augmentation}
\author[A,B,C,\textdagger]{\fnms{Wenda}~\snm{Li} }
\author[A,C,\textdagger]{\fnms{Kaixuan}~\snm{Chen} }
\author[A,C,\textdagger]{\fnms{Shunyu}~\snm{Liu} }
\author[D,E]{\fnms{Tongya}~\snm{Zheng}}
\author[A,C]{\fnms{Wenjie}~\snm{Huang}}
\author[A,C;*]{\fnms{Mingli}~\snm{Song}}
\address[A]{State Key Laboratory of Blockchain and Security, Zhejiang University}
\address[B]{School of Software Technology, Zhejiang University}
\address[C]{Hangzhou High-Tech Zone (Binjiang) Institute of Blockchain and Data Security}
\address[D]{Big Graph Center, School of Computer and Computing Science, Hangzhou City University}
\address[E]{College of Computer Science and Technology, Zhejiang University Hangzhou, China}

\begin{abstract}
Mini-batch Graph Transformer~(MGT), as an emerging graph learning model, has demonstrated significant advantages in semi-supervised node prediction tasks with improved computational efficiency and enhanced model robustness. However, existing methods for processing local information either rely on sampling or simple aggregation, which respectively result in the loss and squashing of critical neighbor information.
Moreover, the limited number of nodes in each mini-batch restricts the model's capacity to capture the global characteristic of the graph. 
In this paper, we propose \textbf{\name}, a novel MGT model that employs a two-stage augmented interaction strategy, transitioning from local to global perspectives, to address the aforementioned bottlenecks.
The \emph{local interaction augmentation}~(LIA) presents a neighbor-target interaction Transformer (NTIformer) to acquire an insightful understanding of the co-interaction patterns between neighbors and the target node, resulting in a locally effective token list that serves as input for the MGT. In contrast, \emph{global interaction augmentation}~(GIA) adopts a cross-attention mechanism to incorporate entire graph prototypes into the target node representation, thereby compensating for the global graph information to ensure a more comprehensive perception. To this end, \name achieves the enhancement of node representations under the MGT paradigm.
Experimental results related to node classification on the ten benchmark datasets demonstrate the effectiveness of the proposed method. 
Our code is available at \url{https://github.com/l-wd/LGMformer}.
\end{abstract}

\end{frontmatter}

\footnotetext[2]{Equal Contribution. Email: \{lwdup, chenkx, liushunyu\}@zju.edu.cn.} 
\footnotetext[1]{Corresponding Author. Email: brooksong@zju.edu.cn.} 
\section{Introduction}

\begin{figure*}[!t]
  \centering
  \includegraphics[width=1.0\linewidth]{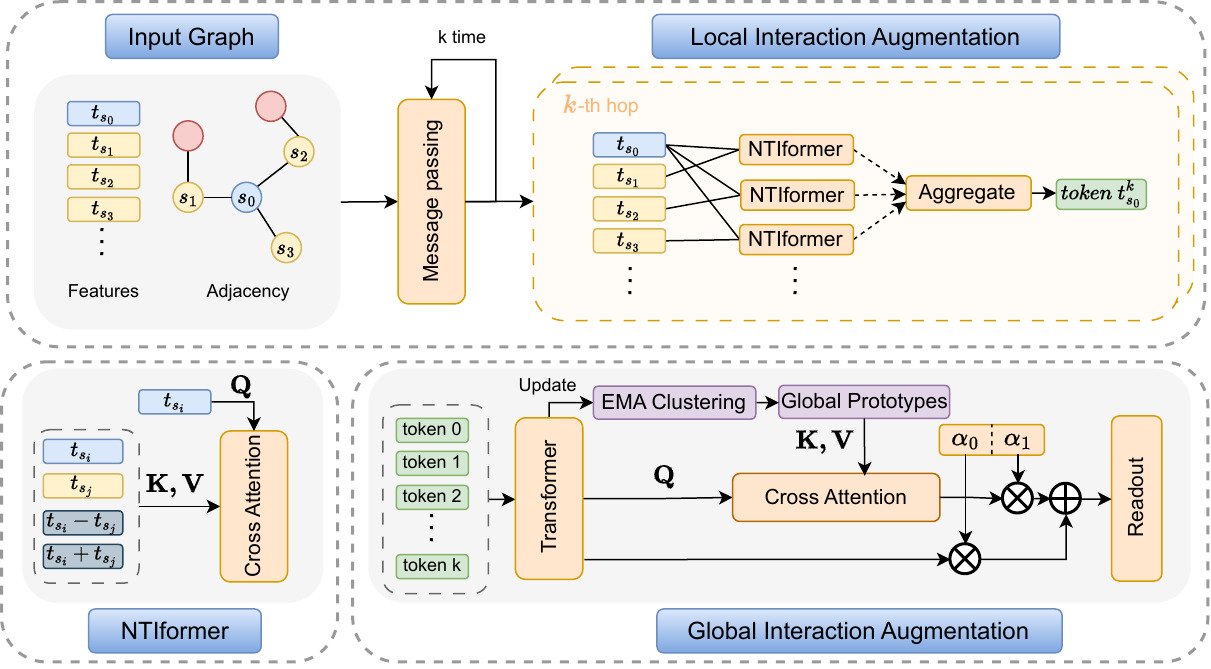}
  \caption{\textbf{Illustration of the proposed \name.} First, \name computes the multi-hop aggregation features for each node in a given graph. Afterwards, during the local interaction augmentation stage, the target node interacts with each neighbor using the NTIformer to obtain high-level semantic information. Finally, in the global interaction augmentation stage, the augmented global information is obtained by interacting with the global prototypes.}
  \label{fig:architecture}
\end{figure*}

Graph Transformer~(GT) models~\cite{rampavsek2022recipe, shirzad2023exphormer, ying2021transformers} apply the Transformer~\cite{vaswani2017attention} architecture into graph-structured data, leveraging the well-established self-attention mechanism to capture the relationship among all graph nodes.
This strategic adaptation not only preserves the advantages of the Transformer model but also mine the inherent graph structure information~\cite{chen2021hybrid,chen2023distribution,hu2024manifold}, 
allowing it to achieve the robust representation learning capabilities and satisfactory prediction performance in various domains, such as academic network analysis~\cite{hao2021walking}, bioinformatics~\cite{ma2023single}, and traffic flow forecasting~\cite{huo2023hierarchical}. 
Therefore, GT models represent a significant advancement in the field of graph representation learning, providing a powerful and adaptable framework for processing and understanding complex graph-structured data.

By taking the scale of the graph data and limitations of computing resources into consideration, GT models can be divided into two categories based on training strategies: Full-batch GT~(FGT)~\cite{dwivedi2020generalization, kreuzer2021rethinking, wu2022nodeformer, ying2021transformers} and Mini-batch GT~(MGT)~\cite{chen2022nagphormer, fu2023vcr, kong2023goat, zhang2022hierarchical, wu2022nodeformer}.  
FGT models employ a straightforward application strategy by treating each node as a token and calculating the global attention between these tokens, allowing for the capture of underlying dependencies among distant nodes.
However, the design of the global attention inherently leads to a quadratic rise in computational complexity as the number of graph nodes increases, accompanied with a substantial demand for computing resources.
Although recent researches~\cite{huang2024tailoring, wu2022nodeformer, wu2024simplifying} effort to reduce algorithmic complexity to linearity using techniques such as the kernelized softmax operator, these methods remain inadequate for effectively training models with extremely large-scale graph data, mainly due to limited computing resources.
To address this challenge, MGT models are developed to optimize computational resource management by strategically dividing the data into smaller, more manageable subsets, and processing them sequentially at each training step.
This approach significantly enhances resource utilization efficiency, especially when dealing with large-scale graph data, which avoids the limitations of traditional full-batch processing methods constrained by computational resources.

Based on the utilization of neighbor information, the existing MGT methods can be classified into three strategies: neighbor sampling~\cite{kong2023goat,zhang2022hierarchical}, subgraph partition~\cite{wu2022nodeformer,wu2023difformer,wu2024simplifying}, and node tokenization~\cite{chen2022nagphormer,fu2023vcr}.
Neighbor sampling and subgraph partition methods,
which involve sampling the $k$ neighbors of target nodes and dividing the graph into separate subgraphs, suffer from the drawback of information loss.
Specifically, neighbor sampling leads to the loss of crucial node information, while subgraph partition negatively impacts the original graph's structure.
Furthermore, node tokenization involves representing each node as a sequence composed of multiple tokens, with each token capturing the neighborhood information of the node at various hop distances. 
However, the neighborhood features are merged into a single vector through a simple aggregation like an average or a summation operation, leading to the squashing of the critical neighborhood.
To this end, we can infer that such MGT models will lead to two following issues: (1) The loss and squashing of critical neighborhood information has a negative impact on graph representation learning. (2) The limited number of nodes in each mini-batch restricts the model's capacity to capture the global characteristic of the graph.

To mitigate the negative impact of the MGT models, we suggest to enhance the integration of critical neighbors' information while compensating for the distant node information from a global perspective. 
The central concept is to alleviate issue of critical node information being squashed without losing the neighborhood information, while expanding the receptive field to a global scale, allowing the model to capture information related to the current node from more distant perspectives.
Thus, the challenge is:
\begin{tcolorbox}[notitle, sharp corners, colframe=darkgrey, colback=white, 
       boxrule=1pt, boxsep=0.5pt, enhanced, 
       shadow={3pt}{-3pt}{0pt}{opacity=1,mygrey},
       title={Challenge},]
\emph{How to design an effective MGT framework that enhances the integration of critical neighbors' information and expands the receptive field to a global scale?}
\end{tcolorbox}

In this paper, we adopt the node tokenization MGT as the basic framework to avoid the issue of losing neighbors' information, and then develop a \textbf{\underline{L}}ocal to \textbf{\underline{G}}lobal interaction augmented \textbf{\underline{M}}ini-batch graph transformer~(\textbf{\name}) to address another two main issues in the MGT paradigm, i.e., the squashing of critical neighbors' information and the absence of a global perspective. Specifically, as shown in Figure~\ref{fig:architecture}, 
we first develop a NTIformer to directly learns the direct interaction patterns between neighboring nodes and the target node. This process is designed to capture the crucial neighboring information associated with the target node into the node token list, thereby enhancing the augmentation of local interaction. Then, 
we employ EMA Clustering to obtain global prototypes capable of representing the entire graph, and utilize cross-attention mechanism achieve effective interaction with the global context. This method aims to compensate for the critical information from distant nodes, thereby achieving the augmentation of global interaction. 
To this end, we have successfully addressed two inherent drawbacks of the MGT via both local and global interaction augmentation.

The contributions of this paper are summarized below:
\begin{itemize}

\item We introduce a two-stage interaction augmentation strategy that progressively enhances the capture of critical nodes from a local to global scale, effectively addressing two predominant issues that exist in the MGT framework. 

\item We design a tailored NTIformer to augment local interaction within the MGT framework, and further achieve global interaction augmentation through cross-attention mechanism between the target nodes and the entire graph prototypes.

\item The proposed \name framework demonstrates its superior performance through extensive experiments conducted on ten node classification datasets, including both homogeneous and heterogeneous graph of varying sizes.

\end{itemize}

\section{Preliminaries}

\noindent
\textbf{Notation.}
Let $G=(V,E)$ represent an undirected and attributed graph, where $V$ is a set of $n$ nodes and $E$ is a set of $m$ edges.
Usually, each node in graph $G$ has a $d$-dimensional node vector $x_{i}$ representing the node's features. Thus graph $G$ has a node feature matrix ${\mathbf{X}}=[x_1, x_2,\dots, x_n]^\top \in \mathbb{R}^{n\times d}$. Edges $E$ describe the relationships between nodes, and all edges form an adjacency matrix $\mathbf{A}$, where $\mathbf{A}_{ij}$ represents the relationship between nodes $v_{i}$ and $v_{j}$.
Let $\mathbf{D}$ be the diagonal degree matrix where $\mathbf{D}_{ii} = \sum_{j} \mathbf{A}_{ij}$. We recall that the normalized adjacency matrix $\hat{\mathbf{A}}$ is defined as $\hat{\mathbf{A}}=\tilde{\mathbf{D}}^{-1 / 2} \tilde{\mathbf{A}} \tilde{\mathbf{D}}^{-1 / 2}$ where $\tilde{\mathbf{A}}$ and $\tilde{\mathbf{D}}$ denote the adjacency matrix and degree matrix with self-loops, respectively.

\noindent
\textbf{Message Passing Scheme.}
The majority of Graph Neural Networks~(GNNs)~\cite{zheng2022transition, chen2023improving} architecture act by propagating information between adjacent nodes of the graph. 
Each message passing GNN~\cite{zhang2020dynamic, zheng2023temporal} aggregates the features of all the nodes around the node and passes them into the next GNN layer or downstream machine learning task~\cite{lee2018graph, liu2024MAM, liu2024PAC} as the new features of the node.
Thus, the Message Passing Scheme at the $\ell$ layer in GNNs can be represented as:
\begin{equation} \label{eq:message_passing_agg}
    b_{i}^{(\ell)} = {\rm Agg}\Big(\Big\{h_j^{(\ell-1)} \mid j \in \mathcal N(i) \Big\}\Big),
\end{equation}
\begin{equation} \label{eq:message_passing_comb}
    h_{i}^{(\ell)} = {\rm Combine}(h_{i}^{(\ell-1)}, b_{i}^{(\ell)}),
\end{equation}
where $\mathcal N(i)$ denotes the adjacent node set of $i$. ${\rm Agg}(\cdot)$ denotes the aggregation operation. ${\rm Combine}(\cdot, \cdot)$ denotes the combination operation.
$h_{i}^{(\ell)}$ is the node features of $i$ at the $\ell$ layer with the initialization of $h_{i}^{(0)}  = x_{i}$ in GNNs.

\noindent
\textbf{Transformer.}
The critical component of Transformer is the self-attention module which computes the correlation between input tokens and extracts important information based on the correlation.
Given a matrix $\mathbf{H} \in \mathbb{R}^{n \times d}$, where $n$ and $d$ represent the number of input tokens and the token feature's dimension, respectively.
A standard self-attention mechanism is formalized as follows:
\begin{equation} \label{eq:attention}
\begin{aligned}
	\mathrm{SelfAttn}(\mathbf{H}) &= \mathrm{softmax}\left(\mathrm{M_{s}}\right)\mathbf{H}\mathbf{W}_V,\\
    \mathrm{M_{s}} &= \frac{\mathbf{H}\mathbf{W}_Q\left(\mathbf{H}\mathbf{W}\right)^\top}{\sqrt{d_K}},
\end{aligned}
\end{equation}
where $\mathbf{W}_Q \in \mathbb{R}^{d \times d_K}$, $\mathbf{W}_K \in \mathbb{R}^{d \times d_K}$, and $\mathbf{W}_V \in \mathbb{R}^{d \times d_V}$ are the projection matrices.

Considering the case of using a sequence $\mathbf{H_1} \in \mathbb{R}^{n_1 \times d_1}$ to query another sequences $\mathbf{H_2} \in \mathbb{R}^{n_2 \times d_2}$, where $n_1$ and $n_2$ are sequence length of $\mathbf{H_1}$ and $\mathbf{H_2}$ respectively, $d_1$ and $d_2$ are hidden dimension of $\mathbf{H_1}$ and $\mathbf{H_2}$ respectively. The standard cross attention is presented as follows: 

\begin{equation} \label{eq:cross_attention}
\begin{aligned}
	\mathrm{CrossAttn}(\mathbf{H_1},\mathbf{H_2}) &= \mathrm{softmax}\left(\mathrm{M_{c}}\right)\mathbf{H_2}\mathbf{W}_V^{\prime} ,\\
    \mathrm{M_{c}} &= \frac{\mathbf{H_1}\mathbf{W}_Q^{\prime}\left(\mathbf{H_2}\mathbf{W}_K^{\prime}\right)^\top}{\sqrt{d_K^{\prime}}},
\end{aligned}
\end{equation}
where $\mathbf{W}_Q^{\prime} \in \mathbb{R}^{d_1 \times d_K^{\prime}}$, $\mathbf{W}_K^{\prime} \in \mathbb{R}^{d_2 \times d_K^{\prime}}$, and $\mathbf{W}_V^{\prime} \in \mathbb{R}^{d_2 \times d_V^{\prime}}$ are the projection matrices.

In the graph transformer, the feature matrix is generally used as an input to the attention mechanism, i.e., $\mathbf{H} = \mathbf{X}$, and then the information interaction between nodes is achieved by following the standard self-attention mechanism.

\section{\name}
In this section, we present the proposed \name, which mainly consists of Local Interaction Augmentation~(LIA) and Global Interaction Augmentation~(GIA).
We first describe the definition of the NTIformer and use it incorporating with the node tokens to enhance the local interaction augmentation. 
Then, we introduce the global prototypes learning implementation based on clustering to achieve global interaction augmentation.
Finally, we give a concrete implementation of the \name and present some details.

\subsection{Local Interaction Augmentation}
In general, detailed local structure and local features are essential, thus any absence or addition of edges or nodes result in a dramatic change of effect.
Currently localized operations such as subgraph partition and neighbor sampling, suffer performance limitations due to information loss caused by inevitable node omission. To avoid information loss, node tokenization operation aggregate information from different hop neighbors as tokens, but leading to the over-squashing of local information.
To adequately exploit the local information, we take a local interaction perspective and analyzing the information between connected nodes.

\begin{observation}\label{thm:observation}
Common information exists between the connected nodes of an arbitrary edge.
Node labels are completely determined by the information contained in node attributes for node classification tasks,
the common information that can significantly enhance the distinctive attributes of target nodes and help in their classification is useful.
\end{observation}
Obviously, the useful common information for node classification should be picked and aggregated as much as possible from the perspective of information interaction, whether the connected nodes of an edge exhibit the same class.

Motivated by the above analysis, we design the neighbor-target interaction transformer~(NTIformer) to extract the useful common information between connected nodes and augment the target representation.
Assuming that the input embedding of local interaction augmented module is $\mathbf{L} = [\mathbf{T}^0, \mathbf{T}^1, \cdots, \mathbf{T}^K]^\top$, where $\mathbf{T}^k \in \mathbb{R}^{n \times d_m}$, $\mathbf{L} \in \mathbb{R}^{(K + 1) \times n \times d_m}$, $K$ is the number of hops, $n$ is the number of nodes, and $d_m$ is the hidden dimension.
Given an arbitrary edge, the connected nodes are $s_i$, $s_j$. 
Consider that the useful parts of the common information for nodes $s_i$ and $s_j$ may exhibit greater disparities due to the class.
We implement synthesis and differentiation effects among neighboring nodes using two distinct operations with regarding $s_i$ as the target node.
The synthesis effect, denoted by $t_{s_i} - t_{s_j}$, contains the information that exclude common information and retain information unique to the target node $s_i$ which may be necessary information for node classification of the target node $s_i$.
The differentiation effect, denoted by $t_{s_i} + t_{s_j}$, contains the information that have enhanced the common information and retain necessary information for node classification of the target node $s_i$.
Next, we regard them as the neighbor-target interaction tokens which are the inputs to the attention mechanism in NTIformer.
In the input embedding $\mathbf{L}$, different hop node embeddings are associated with higher-order information due to multi-hop aggregation.
The neighbor-target interaction tokens of the $k$ hop for target node $s_i$ is calculated as:
\begin{equation} \label{eq:tokens}
    E_{s_i,s_j}^{k} = [t_{s_i}^{k},t_{s_j}^{k},t_{s_i}^{k}-t_{s_j}^{k},t_{s_i}^{k}+t_{s_j}^{k}]^\top,
\end{equation}
where $t_{s_i}^{k}$ and $t_{s_j}^{k}$ are embeddings of nodes $s_i$ and $s_j$ of $k$ hop respectively.

In contrast to the standard transformer, NTIformer avoids the use of a quadratic complexity self-attention mechanism for computing pairwise similarities between tokens. Our motivation lies in enhancing the information pertaining to the target nodes. To achieve this, we consider the target node as a query and employ cross attention to determine the attention between the target node and other tokens as:

\begin{equation} \label{eq:NTIformer}
	\mathrm{NTIformer}(t_{s_i}^k, t_{s_j}^k) = \mathrm{CrossAttn}(t_{s_i}^k, E_{s_i,s_j}^{k}).
\end{equation}

For the target node $s_i$, the common information interaction of a single edge is not enough. 
Thus, to avoid missing any neighboring information and acquire more abundant and complete common information, it is necessary to interact with each of its neighboring nodes and then aggregate the interacted information.
Therefore, the node representation of $k$ hop after common information interaction can be calculated as:
\begin{equation} \label{eq:agg_filter}
\begin{split}
    \mathrm{AN}(t_{s_i}^k) =
    {\rm Agg}\Big(\Big\{\mathrm{NTIformer}(t_{s_i}^k, t_{s_j}^k) \mid s_j \in \mathcal N(s_i) \Big\}\Big).
\end{split}
\end{equation}

Moreover, the high-level semantic information existing in the graph is also essential for target nodes, which originates from local complex structural interactions and nodes features indirectly connected to the target node under the guidance of the graph structure, and important information for target node classification can be captured in the interaction with them.
A straightforward approach is to stack multiple layers of the general-purpose information interactions of neighboring nodes.
However, stacking directly cannot interact well with multi-hop node information which is restricted from propagation by neighboring nodes~\cite{chen2020measuring}, on the one hand, and lacks scalability due to the full-batch processing, on the other hand.
Therefore, we can interact and aggregate with the neighbors by using different hop tokens in $\mathbf{L}$ , which in turn can avoid the above problem.
For the target node $s_i$, we make the node features obtained at each hop interact with its direct neighboring nodes respectively.
Thus, the local interaction augmented token list of the target node ${s_i}$ can be obtained following:
\begin{equation} \label{eq:hop_agg_filter}
    \mathrm{LIA}(\mathbf{L}_{s_i}) = \Big[\mathrm{AN}(t_{s_i}^0), \mathrm{AN}(t_{s_i}^1), \cdots, \mathrm{AN}(t_{s_i}^k)\Big]^\top.
\end{equation}

The node representations obtained in this way not only take into account feature information from nodes of different hop, but also interact with the structure of neighboring nodes and allow us to learn richer node representations in a mini-batch manner.

\subsection{Global Interaction Augmentation}
\begin{algorithm}[t]
  \caption{EMA Clustering}
  \label{alg:global_atten}
  \begin{algorithmic}[1]
      \Require Embedding of batch nodes: $\mathbf{G}$.The batch norm module: $\mathrm{bn}()$.
               Function to find the nearest center for each feature: $\mathrm{FindNearest}(,)$. $\gamma$ is a hyperparameter.
      \Function{UPDATE}{$\mathbf{G}$}
          \State   $\mathbf{G} = \mathrm{bn}(\mathbf{G})$
          \State   $D = \mathrm{FindNearest}(\mathbf{G}, \mathbf{P})$
          \State   $c = c \cdot \gamma + D^T\mathbf{1} \cdot (1 - \gamma)$
          \State   $v = v \cdot \gamma + D^T\mathbf{G} \cdot (1 - \gamma)$
          \State   $v = v/c$
          \State   $\mathbf{P} = v \cdot bn.running\_std+ bn.running\_mean$
      \EndFunction  
  \end{algorithmic}
\end{algorithm}
Global information converge the model's global comprehension over the whole graph, which can provide rich interaction information for specific nodes.
Especially in mini-batch training, since the consideration of information is mostly confined around the target node, local bias can easily be introduced due to local monotonous information, and global features can compensate for the lack of global information on the target node.
Based on the above, we design global interaction with the clustering so as to learn richer and more comprehensive global features.
The clustering method is an operation based on all nodes so that it provides global feature information

\name uses the common K-Means method as a global clustering method and regards the cluster centers as queryable global prototypes~\cite{hao2023heterogeneous}.
Due to the limitation of mini-batch training, providing all nodes for K-Means to learn at once is not possible in training, so we dynamically update the cluster centers of K-Means via the exponential moving average (EMA) algorithm~\cite{kong2023goat}, which is summarized in Algorithm~\ref{alg:global_atten}.
Notably, we remove the additional structural encoding portion as the information provided to the global entity undergoes a more intricate structural interaction based on the local to global paradigm.

Assuming that the input embedding of global interaction augmented module is $\mathbf{G}$, and $\mathbf{P} \in \mathbb{R}^{n_c \times d_c}$ is the learned graph prototypes via K-Means, where $n_c$ is the number of prototypes and $d_c$ is the hidden dimension. 
We update $\mathbf{P}$ each epoch as:
\begin{equation} \label{eq:update_prototypes}
    \mathbf{P} = \mathrm{UPDATE}(\mathbf{G}),
\end{equation}
Global interaction allows each specific node to directly access these prototypes to complement global insights, implemented as follows:
\begin{equation} \label{eq:global_attention}
\begin{split}
    \mathrm{GIA}(g_{s_i}) = \mathrm{CrossAttn}(g_{s_i},\mathbf{P}),
\end{split}
\end{equation}
where $g_{s_i}$ is the embedding of the given node $u$.

\subsection{Implementation Details}
Given graph features $\mathbf{X}$ and normalized adjacency matrix $\hat{\mathbf{A}}$.We use the token list formed by the node features after multi-hop aggregation before training: $\mathbf{U} = [\hat{\mathbf{A}}^{0}\mathbf{X}, \hat{\mathbf{A}}^{1}\mathbf{X}, \cdots, \hat{\mathbf{A}}^{K}\mathbf{X}]^\top$, where $K$ is the number of hops.
Thus, for the target node $s_i$, the input token list is denoted as $\mathbf{U_{s_i}} = [u_{s_i}^0, u_{s_i}^1, \cdots, u_{s_i}^k]^\top$, where $u_{s_i}^0$ is the original node feature without aggregation.
Then we map $\mathbf{U_{s_i}}$ to node embeddings token list in the latent space with a neural layer, i.e., $\mathbf{L_{s_i}} = \boldsymbol{f}_{\theta}(\mathbf{U_{s_i}})$, where $\boldsymbol{f}_{\theta}(\cdot)$ can be a shallow (e.g., one-layer) MLP.

Next, the local interaction augmentation module is applied on $\mathbf{L_{s_i}}$ to get the local augmented embedding token list: $\mathbf{Z_{s_i}^{(0)}} = \mathrm{LIA}(\mathbf{L_{s_i}}) + \mathbf{L_{s_i}}$.
Note that in the local interaction augmentation module, we use independent attention training parameters for aggregated features with different hops.
The token list is then put into the transformer encoder :
\begin{equation} \label{eq:transformer_local}
\begin{aligned}
    \mathbf{Z}^{\prime(\ell)}_{local} &=\mathrm{MHA}\left(\mathrm{LN}\left(\mathbf{Z}^{(\ell-1)}_{s_i}\right)\right)+\mathbf{Z}^{(\ell-1)}_{s_i}, \\
    \mathbf{Z}^{(\ell)}_{local} &=\mathrm{FFN}\left(\mathrm{LN}\left(\mathbf{Z}^{\prime(\ell)}_{local}\right)\right)+\mathbf{Z}^{\prime(\ell)}_{local}, 
\end{aligned}
\end{equation}
where $\ell=1, \ldots, L$ implies the $\ell$-th layer of the transformer encoder. $\mathrm{LN}(\cdot)$ denotes layer normalization. $\mathrm{FFN}(\cdot)$ refers to the feed-forward neural Network.

Meanwhile, global interaction augmentation allow the token list to give attention to the global prototypes obtained through clustering as:
\begin{equation} \label{eq:transformer_global}
\begin{aligned}
      \mathbf{Z}^{(\ell)}_{global} =\mathrm{GIA}(\mathbf{Z}^{(\ell-1)}_{s_i})+\mathbf{Z}^{(\ell-1)}_{s_i}.
\end{aligned}
\end{equation}

\begin{equation} \label{eq:transformer_all}
  \begin{aligned}
        \mathbf{Z}^{(\ell)}_{s_i} = \mathbf{Z}^{(\ell)}_{local} + \mathbf{Z}^{(\ell)}_{global}.
\end{aligned}
\end{equation}
Finally, to obtain the final representation vector of node $u$, readout functions (e.g. mean, sum, attention) will be applied on $\mathbf{Z}^{\prime(\ell)}_{s_i}$.
Figure~\ref{fig:architecture} depicts the architecture of \name.

\textbf{Complexity Analysis.}
The overall time complexity of \name is $O(n(K+1)^2d + (K+1)Ed + n(K+1)n_cd)$, where $K$ is the number of hops, $n$ is the number of nodes, $E$ is the number of edges in graph, $d$ is the hidden dimension, and $n_c$ is the number of cluster centers.

\section{Experiments}
\subsection{Experimental Setup}

\paragraph{Datasets.}
We have conducted experiments on twelve benchmark datasets, including nine small-scale datasets and three relatively large-scale datasets, which contain homophilic and heterophilic datasets respectively.
In the small-scale datasets, computer, photo, and wikics are homophilic graphs, and roman-empire, minesweeper, tolokers, and questions are heterophilic~\cite{lim2021large} graphs.
For computer and photo, we split the dataset into train, valid, and test sets as 60\%:20\%:20\%.
As for the other small-scale datasets, we use the official splits provided in their respective papers~\cite{mernyei2020wiki, platonov2023critical}.
In large-scale datasets, ogbn-arxiv is homophilic dataset from OGB~\cite{hu2020open}. Pokec and twitch-gamer~\cite{lim2021large} are heterophilic datasets.
The split of the large dataset follows the settings of~\cite{hu2020open, lim2021large}. We have adopted the homophily score calculation method from ~\cite{lim2021large}.

\begin{table*}[!t]
  \centering
  \caption{Average results of small-scale homophilic and heterophilic datasets. $\pm$ corresponds to one standard deviation of the average evaluation over 10 trials. Accuracy is reported for roman-empire, computer, photo, and wikics. ROC AUC is reported for minesweeper, tolokers, and questions. Bold and underline indicate the best and second best results, respectively. ``AR'' means average rank of results on small-scale datasets.}
  \vspace{-5pt}
  \setlength{\tabcolsep}{4pt} 
  \label{tab:small_table}
  \renewcommand{\arraystretch}{1.3}
  \begin{tabular}{llcccccccc}
      \toprule
      \multicolumn{2}{c}{\multirow{2}*{Method}} & \multicolumn{4}{c}{Heterophilic} & \multicolumn{3}{c}{Homophilic} & \multirow{2}*{AR}\\
      \cmidrule(lr){3-6} \cmidrule(lr){7-9}
      &                                                                                                         & roman-empire     & minesweeper        & tolokers          & questions         & computer          & photo             & wikics & \\ 
    \midrule
      \multirow{3}*{\begin{tabular}[l]{@{}l@{}}GNNs\end{tabular}} & GCN                                                                           & 73.69 $\pm$ 0.74 & 89.75 $\pm$ 0.52   & 83.64 $\pm$ 0.67  & 76.09 $\pm$ 1.27  & 89.65 $\pm$ 0.52  & 92.70 $\pm$ 0.20  & 77.47 $\pm$ 0.85 & 5.7  \\ 
      & GAT                                                                                               & 53.45 $\pm$ 0.27 & 72.23 $\pm$ 0.56   & 77.22 $\pm$ 0.73  & $\underline{76.28 \pm 0.64}$  & 90.78 $\pm$ 0.13  & 93.87 $\pm$ 0.11  & 76.91 $\pm$ 0.82 & 7.5 \\ 
      & GPRGNN                                                                                            & 64.85 $\pm$ 0.27 & 86.24 $\pm$ 0.61   & 72.94 $\pm$ 0.97  & 55.48 $\pm$ 0.91  & 89.32 $\pm$ 0.29  & 94.49 $\pm$ 0.14  & 78.12 $\pm$ 0.23 & 8.8 \\ 
    \midrule
      \multirow{3}* {\begin{tabular}[l]{@{}l@{}}Scalable\\ GNNs\end{tabular}} & GraphSAINT                 & 68.57 $\pm$ 0.42 & 88.47 $\pm$ 1.34   & 82.08 $\pm$ 0.93  & 68.40 $\pm$ 4.62  & 90.22 $\pm$ 0.15  & 91.72 $\pm$ 0.13  & 65.17 $\pm$ 2.13 & 7.8 \\ 
      ~ & PPRGo                                                                                           & 70.78 $\pm$ 0.51 & 64.88 $\pm$ 1.50   & 70.14 $\pm$ 0.88  & 57.75 $\pm$ 0.81  & 88.69 $\pm$ 0.21  & 93.61 $\pm$ 0.12  & 76.92 $\pm$ 0.72 & 10.2 \\ 
      ~ & GRAND+                                                                                          & 16.50 $\pm$ 0.34 & 68.11 $\pm$ 1.10   & 71.56 $\pm$ 0.65  &  71.29 $\pm$ 1.30 & 88.74 $\pm$ 0.11  & 94.75 $\pm$ 0.12  & 78.10 $\pm$ 0.60
      & 9.5 \\ 
    \midrule
      \multirow{6}*{\begin{tabular}[l]{@{}l@{}}Scalable\\ Graph\\ Transformers\end{tabular}} & GraphGPS   & $\underline{82.00 \pm 0.61}$ & 90.63 $\pm$ 0.67   & $\underline{83.71 \pm 0.48}$  & 71.73 $\pm$ 1.47  & 91.19 $\pm$ 0.54  & 95.06 $\pm$ 0.13  & $\textbf{78.66}$ $\pm$ $\textbf{0.49}$ & 3.7 \\ 
      ~ & DIFFormer                                                                                       & 79.10 $\pm$ 0.32 & $\textbf{90.89}$ $\pm$ $\textbf{0.58}$   & 83.57 $\pm$ 0.68  & 72.15 $\pm$ 1.31  & $\underline{91.99 \pm 0.76}$  & 95.10 $\pm$ 0.47  & 73.46 $\pm$ 0.56  & $\underline{3.2}$ \\ 
      ~ & NodeFormer                                                                                      & 64.49 $\pm$ 0.73 & 86.71 $\pm$ 0.88   & 78.10 $\pm$ 1.03  & 74.27 $\pm$ 1.46  & 86.98 $\pm$ 0.62  & 93.46 $\pm$ 0.35  & 74.73 $\pm$ 0.94 &8.3\\ 
      ~ & GOAT                                                                                            & 71.59 $\pm$ 1.25 & 81.09 $\pm$ 1.02   & 83.11 $\pm$ 1.04  & 75.76 $\pm$ 1.66  & 90.96 $\pm$ 0.90  & 92.96 $\pm$ 1.48  & 77.00 $\pm$ 0.77 &6.5\\ 
       \cmidrule(lr){2-10}
      ~ & NAGphormer                                                                                      & 74.34 $\pm$ 0.77 & 84.19 $\pm$ 0.66   & 78.32 $\pm$ 0.95  & 68.17 $\pm$ 1.53  & 91.22 $\pm$ 0.14  & $\underline{95.49 \pm 0.11}$  & 77.16 $\pm$ 0.72 &5.7\\ 
       \cmidrule(lr){2-10}
      ~ & \name                                                                                           & $\textbf{83.71}$ $\boldsymbol{\pm}$ $\textbf{0.64}$ & $\underline{90.87 \pm 0.44}$ & $\textbf{84.07}$ $\pm$ $\textbf{1.03}$ & $\textbf{77.75}$ $\pm$ $\textbf{1.26}$ & $\textbf{92.08}$ $\pm$ $\textbf{0.25}$ & $\textbf{95.59}$ $\pm$ $\textbf{0.30}$ & $\underline{78.28 \pm 0.69}$ &$\textbf{1.2}$\\ 
      
    \bottomrule
  \end{tabular}
\end{table*}

\paragraph{Baselines.}
We compared our method with 12 advanced baselines,
These include four full-batch GNN methods: GCN~\cite{kipf2017semisupervised}, GAT~\cite{Petar2018graph}, GPRGNN~\cite{chien2020adaptive}, and four scalable GNN methods: GraphSAINT~\cite{zeng2019graphsaint}, PPRGo~\cite{bojchevski2020scaling}, and GRAND+~\cite{feng2022grand+}.
For graph transformer, we compare GraphGPS~\cite{rampavsek2022recipe}, DIFFormer~\cite{wu2023difformer}, NodeFormer~\cite{wu2022nodeformer}, GOAT~\cite{kong2023goat}, and NAGphormer~\cite{chen2022nagphormer}
Among them, GOAT is the method using neighbor sampling to extend large graphs, NodeFormer and DIFFormer are the method using subgraph partition to extend large graphs, and NAGphormer is the tokenized graph transformers.

\begin{table}[!t]
  \centering
  \normalsize
  \caption{Graph Datasets and Statistics.}
  \vspace{-11pt}
  \label{statistics}
  \resizebox{0.48\textwidth}{!}{
  \begin{tabular}[t]{lcrr}
    \toprule
    \textbf{Dataset}   &  \textbf{Homophily Score} &  \textbf{\#Nodes}   & \textbf{\#Edges}    \\ 
    \midrule
    Computer     &  $0.700$ &  $13,752$   &  $245,861$  \\
    Photo     &  $0.772$ &  $7,650$   &  $119,081$  \\
    WikiCS     &  $0.568$ &  $11,701$   &  $216,123$  \\
    \midrule
    roman-empire     &  $0.023$ &  $22,662$   &  $32,927$  \\
    minesweeper     &  $0.009$ &  $10,000$   &  $39,402$  \\
    tolokers     &  $0.187$ &  $11,758$   &  $519,000$  \\
    questions     &  $0.072$ &  $48,921$   &  $153,540$  \\
    \midrule
    ogbn-arxiv  & $0.416$ &  $169,343$   &  $1,166,243$  \\
    twitch-gamers  & $0.090$ &  $168,114$   &  $6,797,557$  \\
    pokec  & $0.000$ &  $1,632,803$   &  $30,622,564$  \\
    \bottomrule
    \end{tabular}
    }
\end{table}
\subsection{Performance on Small-Scale Datasets}
The results presented in Table~\ref{tab:small_table} demonstrate the outstanding performance of our proposed method, \name, across multiple datasets. One of the key advantages of \name over GNN-based models is its ability to capture higher-order semantic information more effectively. This is achieved by allowing different-hop information to flow freely among neighboring nodes, as opposed to constrained message flow based on the graph structure, as observed in models such as GCN and GAT. Compared to sampling-based GNN methods like GraphSAINT and GRAND+ and graph transformer methods like GOAT, the comprehensive inclusion of neighboring nodes enables \name to achieve superior performance by incorporating more local information. Moreover, in comparison to NAGphormer, \name demonstrates significant performance gains, confirming the efficacy of local and global interaction augmentation. Additionally, unlike other approaches that struggle to balance performance on heterophilic graphs, \name facilitates information aggregation among heterophilic nodes through local information interaction augmentation. As a result, it attains competitive performance on heterophilic datasets, particularly the roman-empire dataset, demonstrating an accuracy improvement of 1.71\%. Furthermore, while maintaining balanced performance on heterophilic graphs, \name also achieves competitive performance on homophilic datasets.

\begin{table}[!t]
  \centering
  
  \caption{Average accuracy of large-scale homophilic and heterophilic datasets. The missing results means the training cannot be finished within an acceptable time budget.}
  
  \label{tab:large_table}
  \setlength{\tabcolsep}{5pt} 
  \vspace{-5pt}
  \renewcommand{\arraystretch}{1.2}
  \begin{tabular}{lccccc}
      \toprule
      Method & ogbn-arxiv & pokec & twitch-gamer & AR \\ 
      \midrule
      GraphSAINT & 66.95 $\pm$ 0.18 & 68.99 $\pm$ 0.27 & 61.77 $\pm$ 0.27 & 5.3    \\ 
      PPRGo & 59.54 $\pm$ 0.02 & 60.84 $\pm$ 0.02 & 59.83 $\pm$ 0.02 &7.0\\ 
      GRAND+ & 36.43 $\pm$ 0.00 & 50.76 $\pm$ 0.00 & - &8.0\\ 
      \midrule
      DIFFormer & 69.86 $\pm$ 0.25 & 73.89 $\pm$ 0.35 & 61.22 $\pm$ 0.12& 4.3\\ 
      NodeFormer & 67.19 $\pm$ 0.83 & 71.00 $\pm$ 1.30 & 62.14 $\pm$ 0.17&4.3 \\ 
      GOAT & $\textbf{72.41}$ $\pm$ $\textbf{0.40}$ & 66.37 $\pm$ 0.94 & 62.27 $\pm$ 0.58 &3.3\\ 
      \midrule
      NAGphormer & 70.13 $\pm$ 0.55 & $\underline{76.59 \pm 0.25}$ & $\underline{64.38 \pm 0.04}$ &$\underline{2.3}$\\ 
      \midrule
      \name & $\underline{71.30 \pm 0.16}$ & $\textbf{81.32}$ $\pm$ $\textbf{0.45}$ & $\textbf{64.70}$ $\pm$ $\textbf{0.11}$&$\textbf{1.3}$ \\ 
      \bottomrule
    \end{tabular}
\end{table}

\subsection{Performance on Large-Scale Datasets}
To demonstrate the scalability of our proposed method, \name, we conducted additional experiments on three large-scale graph datasets. In terms of the baseline comparison, we selected three scalable GNN models and four graph transformer methods that offer acceptable computational costs. The results presented in Table~\ref{tab:large_table} indicate that \name performs exceptionally well on large-scale graphs, particularly on Pokec, with an accuracy gain of up to 4.73\%. Comparing \name to NAGphormer, we observe significant performance improvements, further validating the effectiveness of our approach. These results highlight the ability of \name to effectively preserve both local and global information through interaction augmentation, thereby addressing the challenges of node classification in large-scale graphs.

\subsection{Ablation Study}
\begin{figure}[!t]
  \centering
  \includegraphics[width=1.0\linewidth]{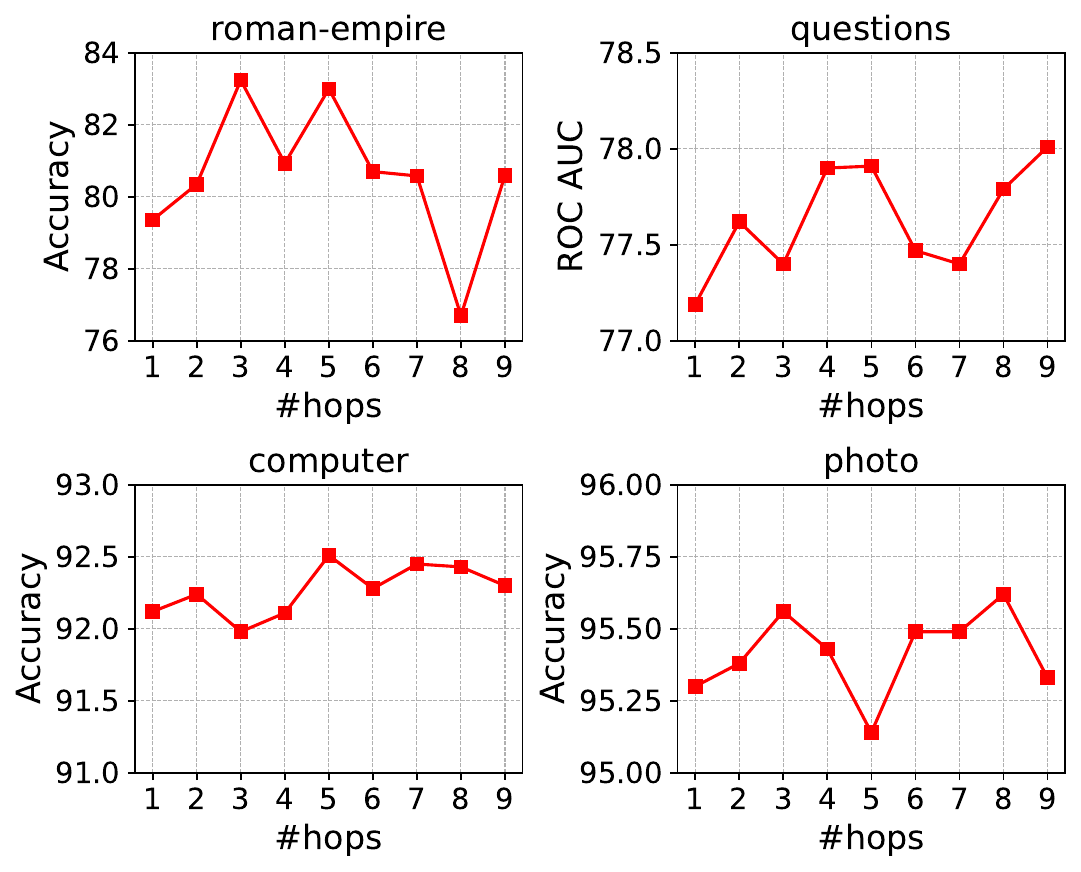}
  \vspace{-0.5cm} 
  \caption{The performance with different length of token list.}
  \label{fig:ablation_hops}
\end{figure}

\paragraph{The length of input token list.}
The length of the input token list determines the number of neighbor hops that the target node can directly access, constituting the local receptive field of the target node. By varying the length of the input token list, we explore the impact of the size of the local receptive field on the node classification performance for the target node. The findings are presented in Figure~\ref{fig:ablation_hops}. We observe that the length of the token list affects performance metrics differently across various datasets, depending on the neighborhood structure of each dataset. While a longer token list can offer more information, it does not necessarily result in improved performance. On the one hand, aggregating more distant information may introduce additional noise. On the other hand, due to the iterative aggregation of tokens from the neighborhood based on the structure, distant tokens from neighboring nodes exhibit higher similarity and thus learn similar representations. Consequently, instead of blindly increasing the length of the input token list, it is valuable to direct attention towards localization and augment the target node information based on local feature and structure interactions.

\begin{table}[!t]
  \centering
  \caption{The performance of different number of global prototypes. ``range'' means the difference between the maximum and minimum results.}
  \label{tab:sizes_table}
  \renewcommand{\arraystretch}{1.2}
  \setlength{\tabcolsep}{5.4pt} 
  \begin{tabular}{lcccc}
      \toprule
      Sizes & roman-empire & questions & computer & photo \\ 
      \midrule
      512 & 83.09 $\pm$ 0.41 & 77.74 $\pm$ 1.36 & 92.38 $\pm$ 0.25 & 95.33 $\pm$ 0.33 \\ 
      1024 & 83.36 $\pm$ 0.36 & 77.71 $\pm$ 1.09 & 92.44 $\pm$ 0.43 & 95.36 $\pm$ 0.36 \\ 
      2048 & 83.06 $\pm$ 0.62 & 77.95 $\pm$ 1.25 & 92.30 $\pm$ 0.44 & 95.36 $\pm$ 0.43 \\ 
      4096 & 83.71 $\pm$ 0.64 & 77.75 $\pm$ 1.26 & 92.08 $\pm$ 0.25 & 95.59 $\pm$ 0.30 \\ \midrule
      range & 0.65 & 0.24 & 0.36 & 0.26 \\ 
      \bottomrule
    \end{tabular}
\end{table}

\begin{figure}[!t]
  \centering
  \includegraphics[width=1.0\linewidth]{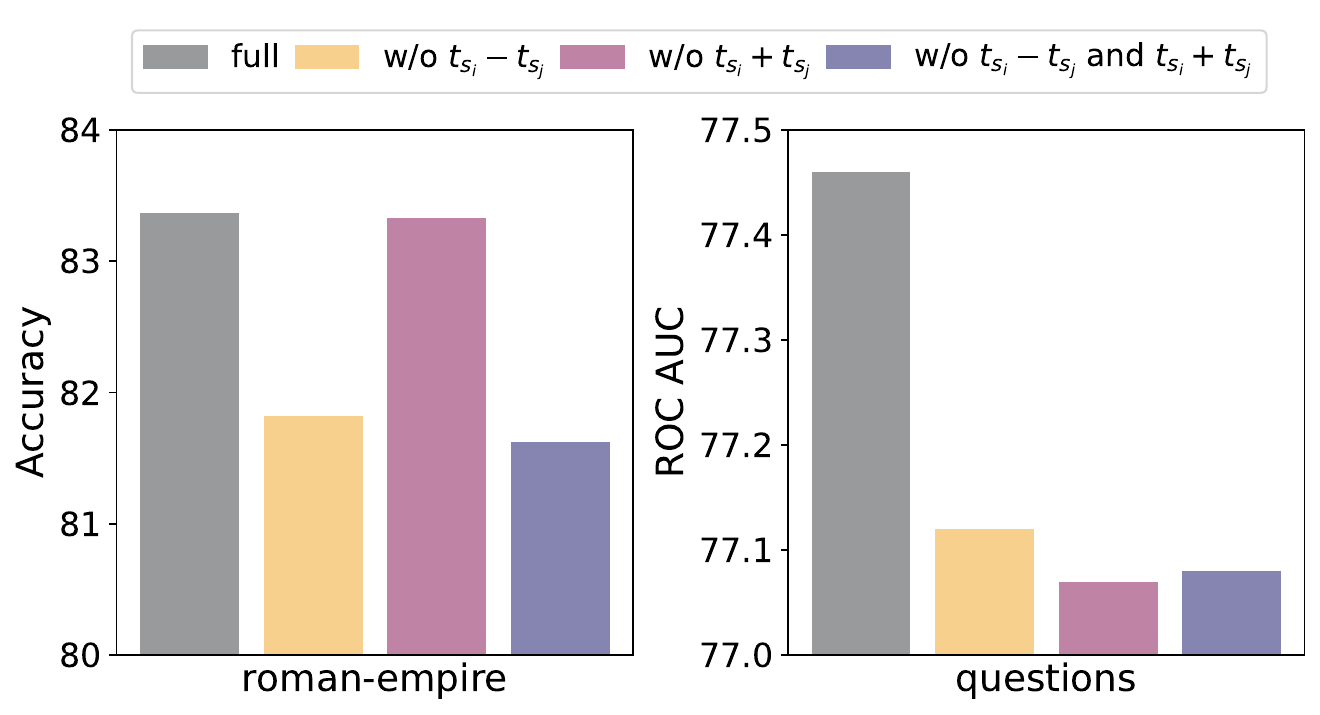}
  \vspace{-0.5cm} 
  \caption{The performance with different token activations in local.}
  \label{fig:ablation_token}
\end{figure}

\begin{figure}[!t]
  \centering
  \includegraphics[width=1\linewidth, height=0.8\linewidth]{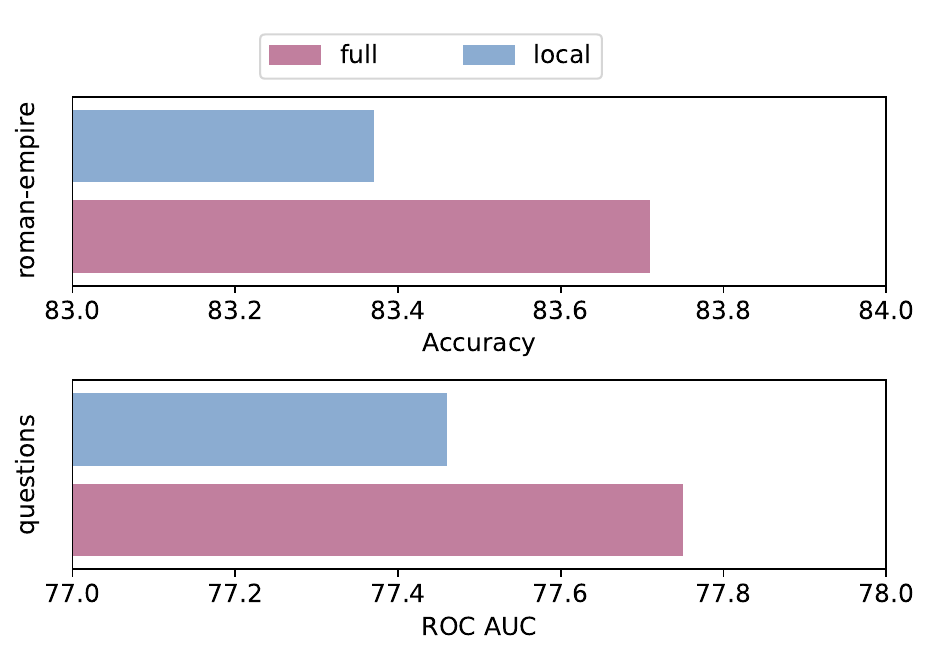}
  \vspace{-0.5cm} 
  \caption{The performance of \name with and without global interaction augmentation for graph datasets.}
  \label{fig:local_global}
\end{figure}

\paragraph{NTIformer tokens.}
We investigate the impact of specific tokens in NTIformer on local interaction augmentation by activating only the designated tokens. The results, as depicted in Figure~\ref{fig:ablation_token}, demonstrate that including these manually constructed tokens leads to performance improvements for graphs such as roman-empire and questions. This suggests that the attention mechanism, which is constructed solely from the nodes' own features, fails to adequately extract the significant common information shared by the nodes on both sides of an edge. On the other hand, the locally constructed tokens foster complex local interactions that efficiently extract crucial information. Thus, our proposed NTIformer module effectively enhances the local interaction module's ability to extract important information more efficiently. Notably, the interaction of tokens $s_i$ and $s_j$ in the roman-empire graph results in a remarkable accuracy gain of up to 1.75\%.

\paragraph{With and without global interaction augmentation.}
We analyze the impact of global interaction augmentation on the performance of \name by comparing the model's performance with and without this augmentation. The results are presented in Figure~\ref{fig:local_global}. It can be observed that the global interaction augmentation module yields a substantial performance improvement for the datasets examined, implying that global information can offer valuable insights to specific nodes, thereby enhancing their performance.

\paragraph{Global number of prototypes.}
We examine the impact of the number of prototypes on global performance within the global interaction augmentation module. The results are presented in Table~\ref{tab:sizes_table}. Although different datasets exhibit varying sensitivities to the number of global prototypes, overall, these variations are minimal. Intuitively, as the number of prototypes increases, the abundance of global information also increases. However, the number of global prototypes has limited influence on the enhancement of the global interaction module in our proposed model, \name. This demonstrates that the purpose of the global module is to compensate for a potential dearth of global information resulting from excessive localization. As a result, a surplus of fine-grained information is not essential at the global level; rather, the model simply necessitates the presence of appropriate global information.

\paragraph{Structural encoding.}
Considering that \name does not currently include an encoding or modeling of the global structure, we aim to investigate the impact of implementing structural encoding in \name. We utilize the eigenvectors of the Laplacian matrix of the graph adjacency matrix, a commonly used approach, as the structural encoding. These eigenvectors are then truncated following the standard practice. To embed the structural encoding, we expand the initial node features in this experiment. To mitigate potential biases introduced by dimension expansion, we employ a comparison method using random values for additional embeddings. The results are presented in Table~\ref{tab:structure_table}. Notably, the influence of structural encoding on \name varies significantly across different datasets, particularly the roman-empire and questions datasets, where incorporating global structural encoding leads to a considerable decrease in model performance. Conversely, the performance decline caused by normally distributed random values is comparatively minor. This discrepancy may arise from the additional bias carried by the structural encoding, which, when embedded into the input features, can impede or even misdirect the augmentation of information interaction in \name.

\begin{table}[!t]
  \centering
  \caption{The performance of with or without structural encoding graph datasets. \name(R) and \name(L) denote using Random values and Laplace structural encoding as additional embedding, respectively}
  \label{tab:structure_table}
  \setlength{\tabcolsep}{2pt}
  \renewcommand{\arraystretch}{1.2}
  \begin{tabular}{lcccc}
      \toprule
      Method & roman-empire & questions & computer & photo \\ 
      \midrule
      \name & 83.71 $\pm$ 0.64 & 77.75 $\pm$ 1.26 & 92.08 $\pm$ 0.25 & 95.59 $\pm$ 0.30 \\ 
      \name(R) & 80.76 $\pm$ 0.72 & 77.53 $\pm$ 1.26 & 92.24 $\pm$ 0.34 & 95.59 $\pm$ 0.33 \\ 
      \name(L) & 79.84 $\pm$ 0.74 & 75.69 $\pm$ 1.21 & 92.30 $\pm$ 0.35 & 95.42 $\pm$ 0.50 \\ 
      \bottomrule
  \end{tabular}
\end{table}

\section{Related Work and Discussion}
\paragraph{Full-batch Graph transformer~(FGT).}
FGT models treat each node as a token and calculating the global attention between these tokens, allowing for the capture of underlying dependencies among distant nodes.
For example, GT~\cite{dwivedi2020generalization} augments the original features of the input nodes by using Laplace feature vectors as position encoding and using them as embeddings and then calculates the global attention between nodes.
NodeFormer~\cite{wu2022nodeformer} employs a kernerlized Gumbel-Softmax~\cite{jang2016categorical} operator to achieve efficient computation, which reduces the algorithmic complexity to linearity.
SGFormer~\cite{wu2024simplifying} designed simple global attention and demonstrates that using a one-layer attention can bring up surprisingly competitive performance.
STAGNN~\cite{huang2024tailoring} uses kernelized softmax to reduce the computational complexity of attention to linear complexity to capture global information and combines with message-passing mechanism to propose subtree attention to capture local information.

\paragraph{Mini-batch Graph Transformer~(MGT).}
The quadratic complexity of the all-pair attention mechanism does not allow the transformer to scale directly to large graphs, and the attention matrix generated by a transformer containing millions of nodes consumes an unacceptably large amount of memory.
Based on the utilization of neighbor information, the existing MGT methods can be classified into three strategies: neighbor sampling, subgraph partition, and node tokenization.

Neighbor sampling obtains the number of nodes within a specified range by sampling from the input perspective.
GOAT~\cite{kong2023goat} uses neighbor sampling to obtain neighbor nodes and calculates the attention between the central node and the neighbor nodes to learn local features, and learns global attention with the help of dimensionality reduction.
ANS-GT~\cite{zhang2022hierarchical} proposes adaptive sampling to integrate four different sampling methods, allowing the model to independently choose the appropriate sampling method and thus the appropriate node to learn local information during training, and introduces the corresponding coarsening graph to learn global information.
Gophormer~\cite{zhao2021gophormer} samples a certain number of self-graphs for each node then uses transformers to learn the node representations on the self-graph, and uses consistent regularization and multi-sample inference strategies to attenuate the uncertainty introduced due to sampling.
Subgraph partition divides the whole graph into different subgraphs and compute all-pair attention on the subgraphs. 
When some full-batch methods such as NodeFormer~\cite{wu2022nodeformer} and SGFormer~\cite{wu2024simplifying} confront a larger graph, the memory they spent is still unacceptable, and they must split the whole graph into subgraphs.
Node tokenization starts from the input and develops a token list based on the target node and train the transformer on the token list.
NAGphformer~\cite{chen2022nagphormer} proposes to pre-extract features from different hop aggregations as token lists and train the transformer directly on it.
PolyFormer~\cite{ma2023polyformer} takes a spectral graph and polynomial perspective to build the token list with stronger representational capabilities, and use a customized polynomial attention mechanism, PolyAttn, to act as a node-based graph filter.


\section{Conclusion}
In this paper, we propose \name, a novel scheme for Mini-batch Graph Transformer based on a two-stage interaction augmentation.
In the local interaction augmentation, we first develop a NTIformer to directly learns the direct interaction patterns between neighboring nodes and the target node. 
Then, in the global interaction augmentation, 
we incorporate the information of global prototypes by interactively cross-attention between local and global representations.
This design paradigm avoids the squashing of critical neighbors' information while compensating for the absence of global information.
Experiments reveal \name's strong performances on both homogeneous and heterogeneous graphs of varying sizes.

\begin{ack}
This work was supported in part by the Joint Funds of the Zhejiang Provincial Natural Science Foundation of China under Grant LHZSD24F020001, in part by the Ningbo Natural Science Foundation under Grant 2023J281, in part by the Zhejiang Province "LingYan" Research and Development Plan Project under Grant 2024C01114, and in part by the Zhejiang Province High-Level Talents Special Support Program "Leading Talent of Technological Innovation of Ten-Thousands Talents Program" under Grant 2022R52046.
\end{ack}
\bibliography{ecai}

\end{document}